\documentclass{article}

\usepackage[preprint]{neurips_2024}

\usepackage[utf8]{inputenc}
\usepackage[T1]{fontenc}    % use 8-bit T1 fonts
\usepackage{booktabs}       % professional-quality tables
\usepackage{subcaption}    % for subfigure environment
\usepackage{amsfonts}       % blackboard math symbols
\usepackage{nicefrac}       % compact symbols for 1/2, etc.
\usepackage{microtype}      % microtypography
\usepackage{xcolor}         % colors
\usepackage{amsmath}
\usepackage{graphicx}
\usepackage{amssymb,bm,amsthm}
\usepackage{subfig}
\usepackage{floatrow}
\usepackage{multirow}
\usepackage{array}
\usepackage{color}
\usepackage{colortbl}
\usepackage{wrapfig}
\usepackage{pythonhighlight}
\usepackage{authblk} % for multiple authors/affiliations
\newcommand\ie{\textit{i.e.,}}
\newcommand\eg{\textit{e.g.,}}

\newcommand{\beq}{\begin{equation}}
\newcommand{\eeq}{\end{equation}}
\newcommand{\beqnn}{\begin{equation*}}
\newcommand{\eeqnn}{\end{equation*}}
\newcommand{\beqy}{\begin{eqnarray}}
\newcommand{\eeqy}{\end{eqnarray}}
\newcommand{\beqynn}{\begin{eqnarray*}}
\newcommand{\eeqynn}{\end{eqnarray*}}
\newcommand{\bit}{\begin{itemize}}
\newcommand{\eit}{\end{itemize}}
\newcommand{\ben}{\begin{enumerate}}
\newcommand{\een}{\end{enumerate}}
\newcommand{\bex}{\begin{example}}
\newcommand{\eex}{\end{example}}
\newcommand{\trace}{\mathrm{trace}}
\newcommand{\balg}[1]{\begin{algorithm} \caption{#1}}
\newcommand{\ealg}{\end{algorithm}}

\newcommand{\balgc}{\begin{algorithmic}[1]}
\newcommand{\ealgc}{\end{algorithmic}}

\newcommand{\bary}{\begin{array}}
\newcommand{\eary}{\end{array}}
\newcommand{\bmx}{\begin{bmatrix}}
\newcommand{\emx}{\end{bmatrix}}
\newcommand{\bsmx}{\left[\begin{smallmatrix}}
\newcommand{\esmx}{\end{smallmatrix}\right]}
\newcommand{\bmxc}[1]{\left[\begin{array}{@{}#1@{}}}
\newcommand{\emxc}{\end{array}\right]}
\newcommand{\bcn}{\begin{center}}
\newcommand{\ecn}{\end{center}}

\newenvironment{theorem}[2][Theorem]{\begin{trivlist}
		\item[\hskip \labelsep {\bfseries #1}\hskip \labelsep {\bfseries #2.}]}{\end{trivlist}}

\title{RL Is Neither a Panacea Nor a Mirage:\\ Understanding Supervised vs. Reinforcement Learning Fine-Tuning for LLMs}
\author{
  Hangzhan Jin\textsuperscript{1,*} \quad
  Sicheng Lv\textsuperscript{2,3} \quad
  Sifan Wu\textsuperscript{2,4} \quad
  Mohammad Hamdaqa\textsuperscript{1} \\
  \textsuperscript{1}PolyTechnique Montreal \quad
  \textsuperscript{2}Mila \quad
  \textsuperscript{3}McGill \quad
  \textsuperscript{4}UDeM
}

\title{RL Is Neither a Panacea Nor a Mirage:\\ Understanding Supervised vs. Reinforcement Learning Fine-Tuning for LLMs}

\begin{document}
\maketitle

\begin{abstract}
Training large language models (LLMs) from scratch is increasingly impractical, making post-training methods such as supervised fine-tuning (SFT) and reinforcement-learning fine-tuning (RL-FT, \eg{} PPO) central to modern practice.  Using an out-of-distribution (OOD) variant of the 24-point card game and new spectrum-based diagnostics, we revisit how these two stages reshape representation of a model and its OOD performance.  Our key findings are:
\begin{itemize}
    \item  Reinforcement-learning fine-tuning can restore most of the OOD performance loss or catastrophic forgetting back to moderate SFT performance level (\eg{} Llama-11B 8.97 \% $\mathbb{R}ightarrow$ 15.38 \%, Qwen-7B 17.09 \% $\mathbb{R}ightarrow$ 19.66 \%). But when SFT pushes the model into a markedly different representation regime: severe overfitting that causes a clear distribution shift — RL-FT can no longer restore the model’s OOD performance.
    \item Unlike previous studies on the singular-value decomposition of the parameter matrices of LLMs, we show that the direction shifts of the singular vectors have much larger impact on the performance of LLMs than the singular values. This holds for both SFT and RL. The shifts concentrate on the directions corresponding to the largest and smallest singular values, leaving the bulk spectrum almost intact, and thus, keeping most intrinsic capacity unchanged.
    \item Low-rank and shallow recovery is surprisingly effective. Restoring the directions of singular vectors corresponding to the top 20 \% of singular values or the first 25 \% of layers recovers 70 to 80 \% of a model’s OOD performance.
    \item Better in, better out.  The stronger the SFT checkpoint is, the more faithfully the OOD ability can be rescued by RL.  Conversely, highly overfitted SFT checkpoints are harder to be pulled back even with aggressive UV restoration.
\end{itemize}
These results reconcile prior reports of RL superior OOD performance: RL primarily counteracts SFT-induced directional drift to reduce the catastrophic forgetting rather than discovering fundamentally new solutions. Our spectrum-aware analysis highlights inexpensive recovery knobs, \ie{} low-rank UV merging and shallow-layer resets, that practitioners can employ before resorting to costly RL fine-tuning. 
\end{abstract}

\section{Introduction}
\label{sec:introduction}
Supervised Fine-Tuning (SFT) is the mostly used method for the post-training of Large Language Models (LLMs). It is recently found that the use of Reinforcement Learning (RL) fine-tuning after SFT can achieve much better performance on complex reasoning tasks, such as symbolic math reasoning~\cite{deepseekai2025deepseekr1incentivizingreasoningcapability, xaigrok2025}, code generation~\cite{mirzadeh2024gsm,jiang2024survey, anthropicclaude2025},  embodied tasks~\cite{lin2025evolvenav, li2025perception}, video prediction~\cite{shi2025enhancing} and etc. 
Such a two-stage fine-tuning paradigm has rapidly become popular because of its advantages over the one-stage SFT~\cite{openr1, wang2025reinforcement}. 

%Numerous studies have been conducted to verify and investigate how RL helps SFT in fine-tuning. 
Numerous studies have investigated how RL helps SFT in post-training: a growing body of work argues that SFT tends to memorize training distributions whereas RL yields better out-of-distribution (OOD) generalization~\cite{kirk2023understanding, chu2025sftmemorizesrlgeneralizes}; others emphasize that KL-regularized RL counteracts SFT's drift from the base model~\cite{fu2025srft}, and that rule-based or structure-aware RL can significantly strengthen reasoning~\cite{xie2025logicrlunleashingllmreasoning}. While these findings established the high-level picture, they were largely \emph{observational} and lacked a deep, mechanistic account of the parameter-level dynamics. \cite{xie2025logicrlunleashingllmreasoning} noted that SFT pulls a model's policy away from its base initialization and that specific RL recipes can boost reasoning, but these insights remained correlational and offered little actionable guidance on balancing the two stages. 

To fill the above gaps, we perform full-parameter SFT and RL finetuning on two popular open models: Llama-3.2-11B~\cite{grattafiori2024llama} and Qwen-2.5-7B~\cite{qwen2.5}. By continuously tracking both in-distribution (ID) and OOD performance on the \textsc{GeneralPoints} card-game benchmark~\cite{zhu2023large, ye2024physics}, a controlled probe of arithmetic reasoning and generalization.%
\footnote{Throughout, \emph{ID} refers to performance on the target task family used for SFT, while \emph{OOD} refers to pretraining-acquired general abilities not explicitly targeted during SFT.}
This controlled environment allows us to disentangle the evolution of memorization, generalization, and alignment across the entire two-stage post-training pipeline.
%we conduct a fine-grained analysis about the relation between SFT and RL stages, explicitly tracking how model performance evolve throughout the two phases. More specifically, we evaluate in-distribution (ID) vs. OOD performance \footnote{Note that, in this paper, ID performance refers to the model ability on the targeted task, \ie{} fine-tuning or training data. OOD performance means the general abilities of LLM that is acquired in pretraining without further fine-tuning.} on the GeneralPoints card game benchmark~\cite{zhu2023large, ye2024physics}, a task designed to probe arithmetic reasoning and generalization in foundation models. We systematically record the generalization behavior of two popular LLMs, Llama-3.2-11B \cite{grattafiori2024llama} and Qwen-2.5-7B \cite{qwen2.5} and how their performance evolves across different training stages. 

%Through experiments, we reveal that \textbf{peak generalization \sitao{peak ood or id?} is achieved early in the SFT phase, only to degrade as SFT continues. The subsequent RL fine-tuning stage substantially mitigates this decay}, recovering up to 85\% of the highest out-of-distribution performance for Llama-3.2-11B-Base and 99\% for Qwen-2.5-7B-Base. 
%Through our analysis of the generalization ability of two-stage fine-tuning, we discovered that the primary role of the RL stage does not enable fundamentally better solutions to OOD tasks, but rather to restore general capabilities compromised during the preceding SFT stage.

Through our experiments, we found a consistent pattern across both models. First, OOD generalization \emph{peaks early} during SFT and then degrades as SFT continues, even while ID performance keeps improving. Second, the subsequent RL stage \emph{substantially mitigates} this OOD decay: it restores up to 99\% of the OOD performance lost during SFT for Qwen-2.5-7B and up to 85\% for Llama-3.2-11B, all while maintaining strong ID competence. Third, this restoration has a \emph{limit}: if SFT is prolonged beyond a threshold, the lost ability becomes only partially recoverable---RL can no longer fully heal the forgetting. Overall, these results suggest that RL does not endow fundamentally new capabilities; instead, it plays a \textbf{restoration} role that recovers general competencies compromised by aggressive SFT.

To uncover the mechanisms, we analyze the model's weight matrices through spectral analysis, specifically, singular-value decomposition (SVD). Recent work has shown that the spectrum of parameter matrices offers an interpretable window onto how its internal representations evolve and how they relate to downstream performance~\cite{staats2025smallsingularvaluesmatter, yunis2024approaching}. With this lens, we track geometric shifts in parameter space across the SFT phase and the subsequent RL phase~\cite{aghajanyan2020intrinsic, yunis2024approaching}.
Our results reveal that the model's intrinsic capacity, which is reflected in the bulk of the spectral density, remains essentially constant throughout training. Instead, OOD performance degradation and its later recovery correlate almost entirely with rotations of the singular vectors at the spectrum's extremes, while the corresponding singular values change little. Thus, contrary to earlier work that emphasized the absolute size of singular values~\cite{staats2025smallsingularvaluesmatter}, our findings indicate that directional shifts in the extremal components, rather than magnitude shifts, govern model's performance.

In summary, our main contributions are:
\begin{itemize}
    \item %Through dynamic analyzing OOD performance of two-stage post-training SFT and RL continually on Llama-3.2-11B and Qwen-2.5-7B, 
    We creatively observe that in two-stage post-training, RL primarily serves as memory restoration rather than creation capability of OOD generalizatio.
    \item Through dynamic analyzing OOD performance for different training extend of SFT, we found that if SFT trains for too long, RL cannot recover the lost ability.
    \item Through spectral analysis under full-parameter optimization, we find that singular values remain stable while OOD forgetting and recovery are driven by directions. %We find that model's capcity keeps stable during SFT and RL through spectral analysis on model weight matrices.
    %\item We find that OOD performance change is caused by the rotation of singular vector. 
\end{itemize}

\section{Preliminaries}
\label{sec:preliminaries}
\subsection{Supervised Fine-Tuning (SFT)}
Supervised Fine-Tuning adapts a pre-trained model to a specific task using a labeled dataset $\mathcal{D} = \{(x_i, y_i)\}$. The standard objective is to minimize the negative log-likelihood (NLL) of the target outputs given the inputs:
\[
    \mathcal{L}_{\text{SFT}}(\theta) = -\sum_{(x_i, y_i) \in \mathcal{D}} \log p_\theta(y_i \mid x_i)
\]

%The gradient derived from this objective exclusively rewards the imitation of the ground-truth outputs $y_i$. While effective for task-specific adaptation, this can force the model's parameters $\theta$ into a narrow region of the loss landscape. Such aggressive optimization on a potentially small or specialized dataset creates the risk of \textbf{catastrophic forgetting}, where the model loses the robust, general-purpose representations acquired during pre-training.

\subsection{Reinforcement Learning (RL) Fine-Tuning}
In contrast to SFT, RL fine-tunes the language model (the policy $\pi_\theta$) by optimizing with a scalar reward signal. The general objective is to maximize the expected reward of the  actions made by the model
\[
    \max_\theta \;\; \mathbb{E}_{x \sim \pi_\theta}[R(x)]
\]
Here, the reward function $R(x)$ evaluates the quality of an action $x$ based on desired attributes like correctness~\cite{ouyang2022training}, clarity~\cite{wang2023pandalm}, or adherence to rules~\cite{bai2022constitutional}. 

In this paper, we employ \textbf{Proximal Policy Optimization (PPO)}~\cite{schulman2017ppo}, a popular RL algorithm that stabilizes training by optimizing a clipped surrogate objective. The PPO objective is:
$$
    L_{\text{PPO}}(\theta) = \mathbb{E}_{t}\left[\min\left(r_t(\theta) A_t, \text{clip}(r_t(\theta), 1 - \epsilon, 1 + \epsilon) A_t\right)\right]
$$
where $r_t(\theta) = \frac{\pi_\theta(a_t|s_t)}{\pi_{\theta_{\text{old}}}(a_t|s_t)}$ is the probability ratio, $A_t$ is the advantage estimate(the excess of an action's return over the value baseline), and $\epsilon$ is a hyperparameter that constrains the policy update step. %By using a reward signal that encourages exploration, RL can guide the model out of the hyperspecialized state induced by SFT and help it rediscover more generalizable parameter configurations, thereby \textbf{restoring} compromised capabilities.
%GRPO, DAPO, GSPO, etc.
\subsection{Spectral Analysis via Singular Value Decomposition (SVD)}
%To analyze the internal mechanisms of fine-tuning, we examine the weight matrices of the model using Singular Value Decomposition (SVD).
For any weight matrix $M \in \mathbb{R}^{m \times n}$, its SVD is given by:
\[
    M = U \Sigma V^\top
\]
where $U \in \mathbb{R}^{m \times m}$ and $V \in \mathbb{R}^{n \times n}$ are orthogonal matrices whose columns are the left and right singular vectors, respectively. $\Sigma \in \mathbb{R}^{m \times n}$ is a rectangular diagonal matrix containing the non-negative singular values, $\sigma_1 \ge \sigma_2 \ge \dots \ge 0$.

In the context of a neural network, the singular values $\{\sigma_i\}$ are often interpreted as the importance of different representational modes, while the singular vectors (the columns of $U$ and $V$) define the directions of these modes. SVD on parameter matrices can help us to analyze the internal mechanisms of fine-tuning and to test a our main hypothesis, \ie{} the generalization drop during SFT is primarily caused by a \textbf{shift of the key singular directions}, not a collapse in the singular values. %SVD provides a formal framework to track the geometric alignment of these directions and to empirically validate that RL's restorative effect corresponds to realigning these critical vectors.
%To analyze the internal mechanisms of fine-tuning, we examine the weight matrices of the model using Singular Value Decomposition (SVD).
\subsection{Principal Angles Between Subspaces}
\label{sec:rotation}
Given the parameter matrix of a base model \(W_{\text{base}}\in\mathbb{R}^{m\times n}\) and its fine-tuned counterpart \(W_{\text{tgt}}\), we quantify the amount of rotations between two subspaces by how much their dominant singular directions have \emph{rotated}, which is a commonly used method in machine learning numerical computation   The computation proceeds in three concise steps.

\paragraph{(i) SVD.}
For each matrix, we keep all singular vectors in our experiments,
\[
  W
  = U\,\Sigma\,V^{\top},
  \qquad
  U\in\mathbb{R}^{m\times r},\;
  V\in\mathbb{R}^{n\times r},
  \tag{2.4.1}
\]
where the columns of \(U\) and \(V\) are orthonormal and
\(\Sigma=\operatorname{diag}(\sigma_1,\dots,\sigma_r)\) with
\(\sigma_1\!\ge\!\dots\!\ge\!\sigma_r\!\ge\!0\), $r$ is the rank($r=\min(m,n)$).
The subsequent computations \(O(\min\{m,n\}^3)\).

\paragraph{(ii) Computation of Principal Angles Between Subspaces (PABS).}
Let \(U_{\text{base}},U_{\text{tgt}}\in\mathbb{R}^{m\times k}\) be the left
singular blocks from the previous step.  Define
\(\!M := U_{\text{base}}^{\!\top}U_{\text{tgt}}\in\mathbb{R}^{r\times r}\).
Since both of them are orthonormal, the singular values of
\(M\) lie in \([-1,1]\) \cite{bjorck1973numerical}.  Suppose the SVD of $M$ is 
\[
  M = {U}_M\,\operatorname{diag}(s_1,\dots,s_r)\,{V}_M^{\!\top},
\]
the \emph{principal angles}
\(\theta_i\in[0,\pi/2]\) between $U_{\text{base}}^{\!\top}$ and $U_{\text{tgt}}$ are
\begin{equation}
  \theta_i \;=\; \arccos(s_i), 
  \quad i=1,\dots,r.
  \tag{2.4.2}
\end{equation}
An identical procedure on
\(V_{\text{base}},V_{\text{tgt}}\) yields angles for the right subspaces.
In practice we clamp the numerical values of \(s_i\) to \([-1,1]\) before
calling \(\arccos\) to avoid floating-point overflow. The principal angles mean the degree to which two parameter matrices are different from each other in terms of singular vectors under the rank r.

\medskip
The angle set \(\{\theta_i\}\) serves as a fine-grained measure of subspace
rotation: \(\theta_i=0\) means the \(i\)-th principal direction is
preserved, whereas values approaching \(\pi/2\) indicate maximal
misalignment.

\section{Evaluation and Analysis}
\label{sec:evaluation_analysis}
To investigate the evolution of OOD vs. ID generalization ability during different post-training stages, we evaluate the finetuned models on the GeneralPoints task~\cite{chu2025sftmemorizesrlgeneralizes} and its OOD variantions. We will introduce the experimental settings, empirical results and analysis in this section.

\subsection{Evaluation Settings}

\textbf{GeneralPoints Game}
The \textit{GeneralPoints} environment~\cite{chu2025sftmemorizesrlgeneralizes} is designed to evaluate the arithmetic reasoning ability of models, which is instantiated on top of \textit{Points24} environment~\cite{zhai2024fine}. Each state $s$ of the environment contains 4 poker cards, described as text directly. The goal is to produce an equation that equals a target number (24 by default) with 4 numbers from the cards used only once. Particularly, the cards $'J', 'Q',$ and $'K'$ are all interpreted  as the same number $10$ in the raw settings. For example, provided with four cards [5,4,10,7], we aim to output the equation \textit{(7-5)*10+4} as the satisfied output. Detailed examples for the GeneralPoints state-action transitions are provided in Appendix~\ref{app:examples}.

\textbf{Out-of-distribution Variation for GeneralPoints Game} To disentangle the learning of format alignment and real arithmetic reasoning ability

To study whether RL really improves the pretrained ability of LLM and how does such ability evolves, we use the evaluation on the variation of GeneralPoints to test the OOD performance of LLM following~\cite{chu2025sftmemorizesrlgeneralizes}. It interprets the symbols $'J', 'Q',$ and $'K'$ all as the same number $10$ and evaluate the model by interpret the symbols as $11, 12,$ and $13$ respectively. In our paper, the input prompt is only text.%, as illustrated in Figure***. 
Through evaluting the model on variation rule, we can have a rigorous evaluation of the generalization ability of models.

\textbf{Models and Setup} %To present experiments that investigate the generalization abilities dynamicly changed with post-training, we 
We use two most popular open-source base models Llama-3.2-11B~\cite{grattafiori2024llama} and Qwen-2.5-8B~\cite{qwen2.5} as the backbone models. Following the standard pipeline of reasoning model post-training~\cite{deepseekai2025deepseekr1incentivizingreasoningcapability}, we first warm-up the model with SFT, and then run RL on top of SFT checkpoint(1100 checkpoint for Llama-3.2-11B and 800 ckpt for Qwen-2.5-8B). The format of the prompt is the same as~\cite{chu2025sftmemorizesrlgeneralizes}. Detailed experimental setting for SFT and RL are in Appendix~\ref{app:setting}.

\begin{figure}[t]        % h = “here”, t = top, b = bottom, p = page of floats
  % --- first subfigure ------------------------------------
  \begin{subfigure}[t]{0.48\linewidth}
    \centering
    \includegraphics[width=\linewidth]{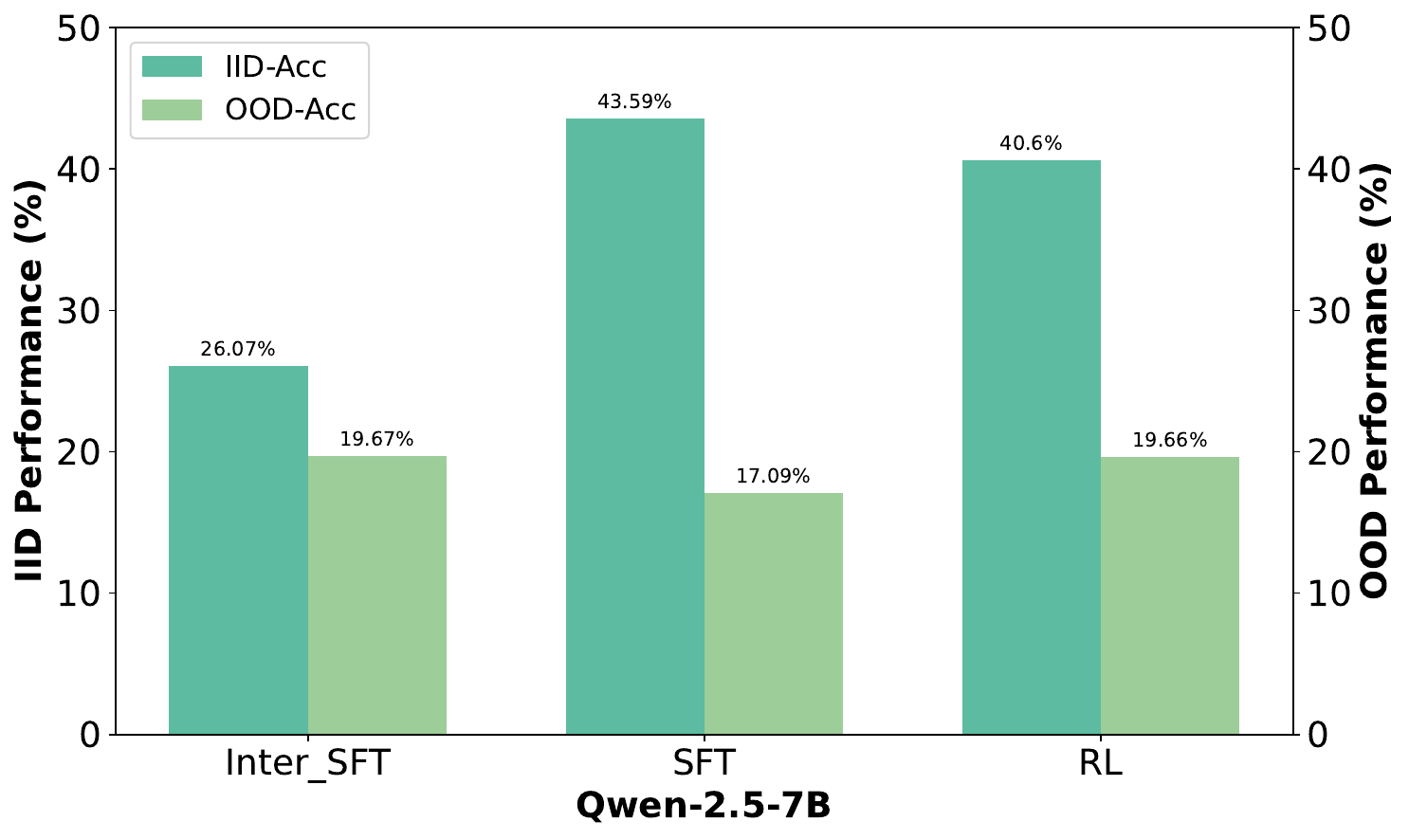} % first-image.pdf/png/eps…
    %\caption{First image caption}
    \label{fig:first}
  \end{subfigure}
  \hfill                           % horizontal space between subfigures
  % --- second subfigure -----------------------------------
  \begin{subfigure}[t]{0.48\linewidth}
    \centering
    \includegraphics[width=\linewidth]{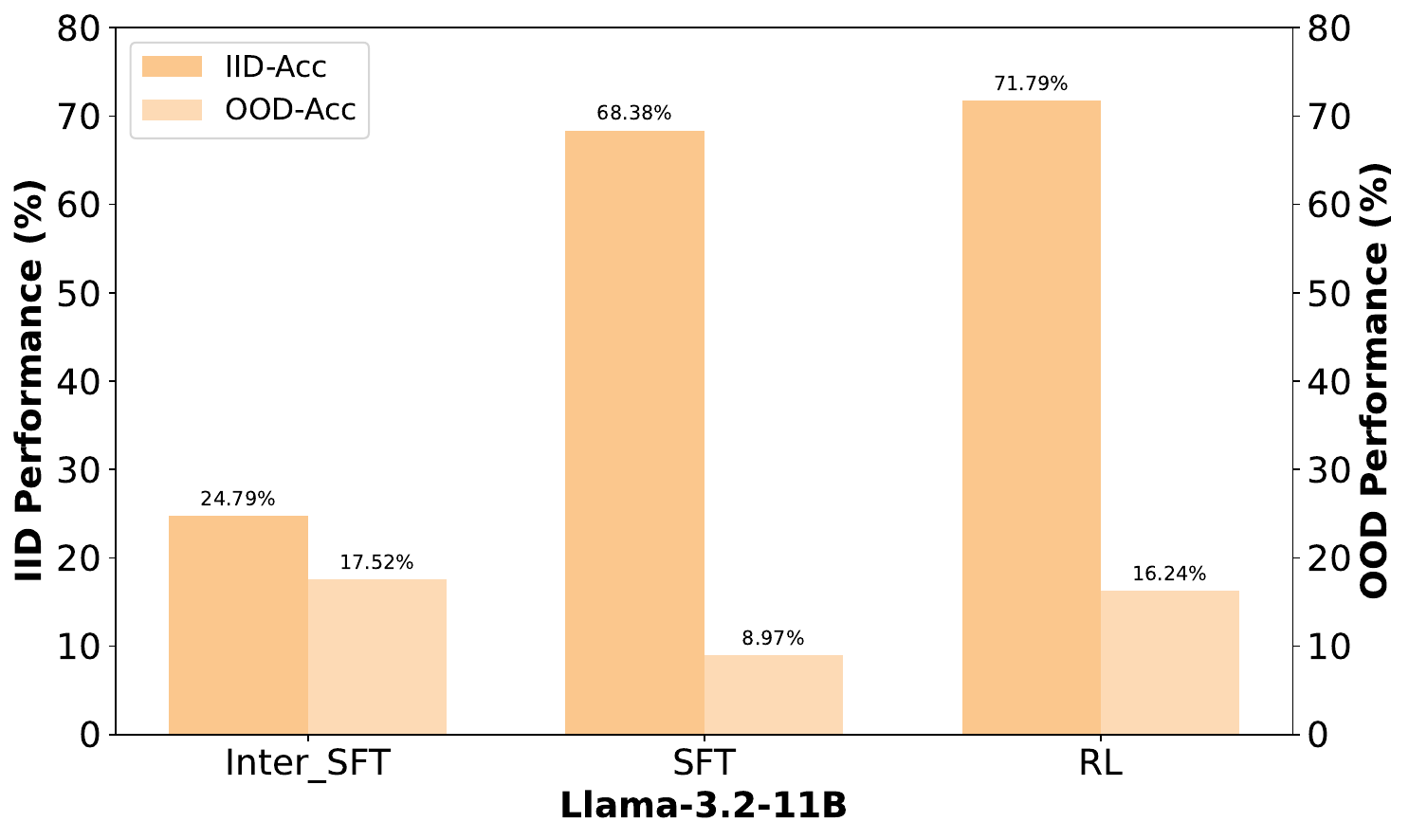}
    %\caption{Second image caption}
    \label{fig:second}
  \end{subfigure}
  \caption{Success rate comparison across checkpoints for Qwen-2.5-7B and Llama-3.2-11B.}
  \label{fig:rl_sft_acc_comp}
\end{figure}

\subsection{Results and Analysis}
\textbf{SFT forgets, RL recovers.} 
As illustrated in Figure \ref{fig:rl_sft_acc_comp}, our experiments reveal the dynamical evolution of the relationship between model specialization on target tasks and forgetting. 

For the two base models, 

their OOD performance first peaks at an Inter\_SFT checkpoint and then declines with continued SFT. The Inter\_SFT checkpoint represents an early-stage snapshot of the model during fine-tuning (e.g., at the 20\% training mark), which we empirically selected by evaluating checkpoints at regular intervals and identifying the one that achieved the highest OOD performance. The subsequent performance drop with further SFT is a phenomenon indicative of "SFT forgetting," where the model begins to overtrained (shrinks) to the in-distribution data, losing its ability to generalize.
For the Llama-3.2-11B model, OOD performance initially rose to a peak of 17.52\% with Inter\_SFT before dropping sharply to 8.97\% after full SFT, dropping 48\% on generalization. Similarly, on the Qwen-2.5-7B model, performance peaked at 19.67\% with Inter\_SFT and subsequently drops 13\% on OOD for full SFT.
The successive RL recover the generalization for 99\% for Qwen-2.5-7B model and 85\% for Llama-3.2-11B. 
In contrast with OOD performance, in-distribution (ID) accuracy increases steadily throughout the SFT process, reaching its maximum value after full SFT as the model becomes highly specialized. However, the subsequent RL phase, which restores OOD generalization, does so by sacrificing an amount of this peak ID performance. 

Overall, SFT's primary function is to specialize a model for ID tasks, but prolonged training degrades OOD performance. RL then serves as a crucial corrective step, not by generating new knowledge, but by recovering the OOD generalization lost during SFT. This highlights a fundamental trade-off, where RL re-balances the model, restoring robust generalization at the cost of a slight reduction in its highly specialized ID accuracy. [hangzhan: TODO, the same, not small]
\begin{figure}[t]
  % --- first subfigure ------------------------------------
  \begin{subfigure}[t]{0.49\linewidth}
    \centering
    \includegraphics[width=\linewidth]{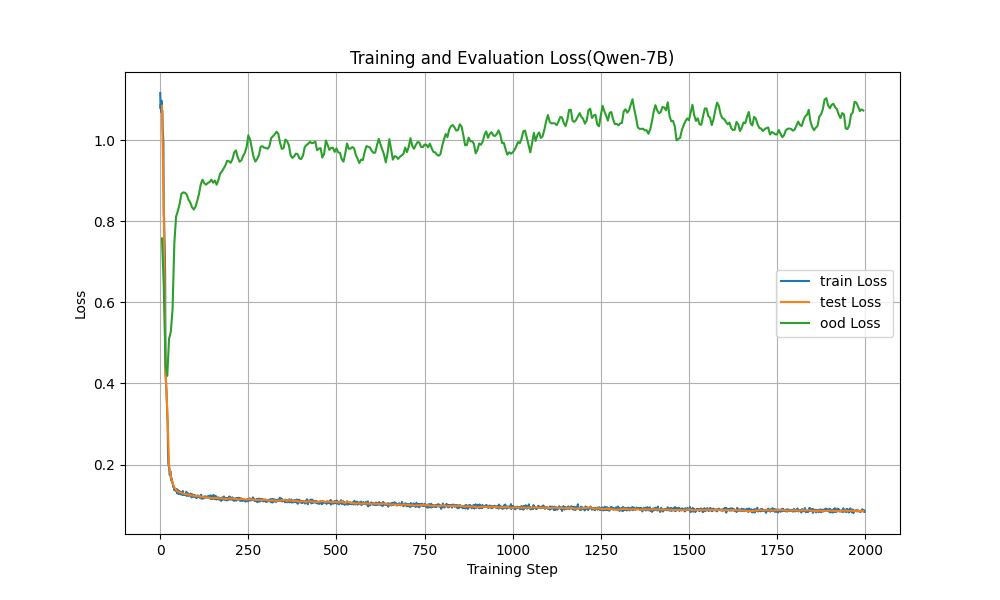}
    \label{fig:first}
  \end{subfigure}
  \hfill                           % horizontal space between subfigures
  \begin{subfigure}[t]{0.49\linewidth}
    \centering
    \includegraphics[width=\linewidth]{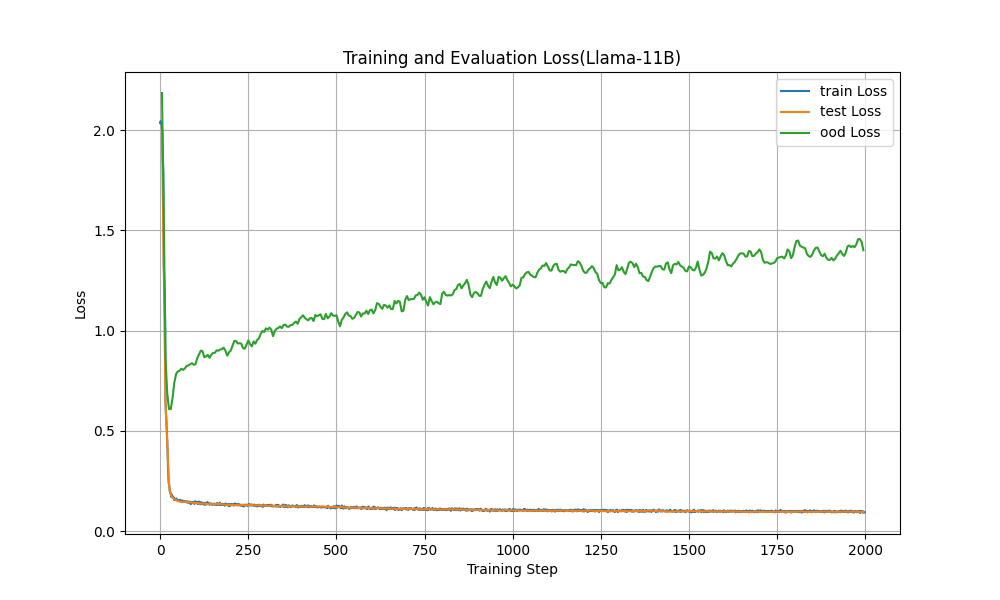}
    %\caption{Second image caption}
    \label{fig:second}
  \end{subfigure}
  \caption{In-distribution(train and test) loss and OOD loss curves for Qwen-2.5-7B and Llama-3.2-11B during SFT.}
  \label{fig:sft_loss}
\end{figure}

\textbf{Monotonic ID improvement and early OOD gain then steady erosion during SFT\footnote{Since the evaluation of accuracy need fulls full rollout in the output, which is slow and computationally expensive, we evaluate the training process by loss, which is more efficient.}.}
As Figure \ref{fig:sft_loss} illustrates, both Qwen-2.5-7B and Llama-3.2-11B show a smooth, monotonic decline in ID train and test loss throughout SFT.
In contrast, their OOD loss falls sharply for the first 100 steps and reachs minima of 0.45 (Qwen) and 0.67 (Llama). Then OOD loss increasing almost linearly and end even higher than where it began.
This pattern is consistent with the performance results from Figure~\ref{fig:rl_sft_acc_comp}:ID accuracy keeps rising while OOD accuracy peaks early and deteriorates thereafter. We suspect that the  later SFT updates reallocate model capacity toward increasingly domain-specific directions, improving in-distribution fit at the expense of generalization, which is a classic manifestation of catastrophic forgetting.

[hangzhan: TODO, For the singular value curves that override each other, we may need to emphasize this phenomenon that indicates the value doesn't change that much after fine-tuning. In addition, we should explain that we recovered the singular value and kept the singular vectors, the OOD and Ind capabilities didn't change after this recovery, this means the singular values have nothing to do with the performance of a model after fine-tuning.]
\begin{figure}[ht]
  \centering
  % ---- 1st row -------------------------------------------------
  \begin{subfigure}[b]{0.32\textwidth}
    \includegraphics[width=\linewidth]{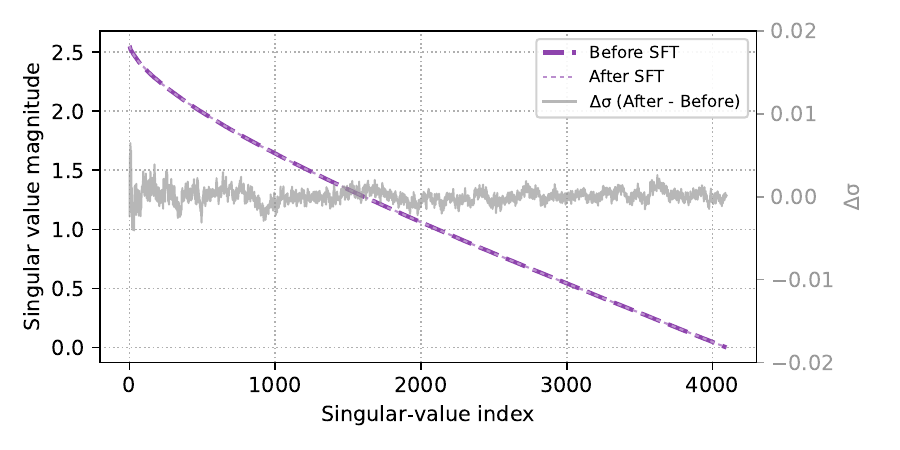}
    \caption{$W_q$ changes during SFT}
    \label{fig:1a}
  \end{subfigure}\hfill
  \begin{subfigure}[b]{0.32\textwidth}
    \includegraphics[width=\linewidth]{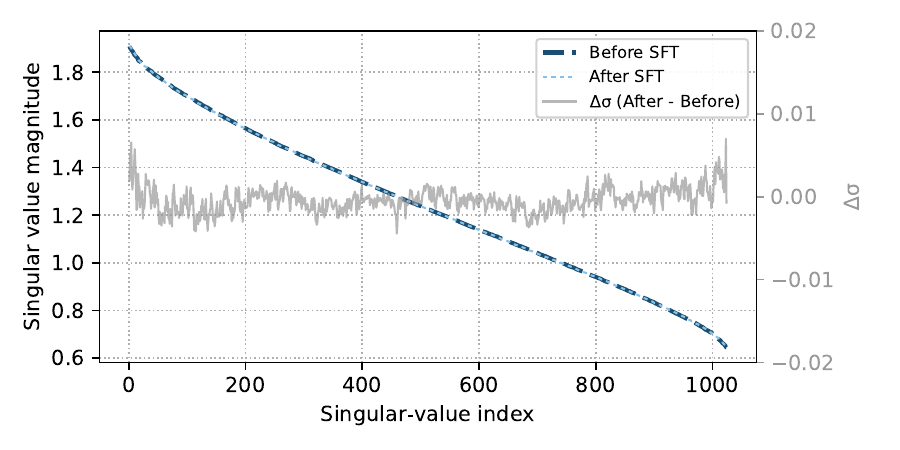}
    \caption{$W_k$ changes during SFT}
    \label{fig:1b}
  \end{subfigure}\hfill
  \begin{subfigure}[b]{0.32\textwidth}
    \includegraphics[width=\linewidth]{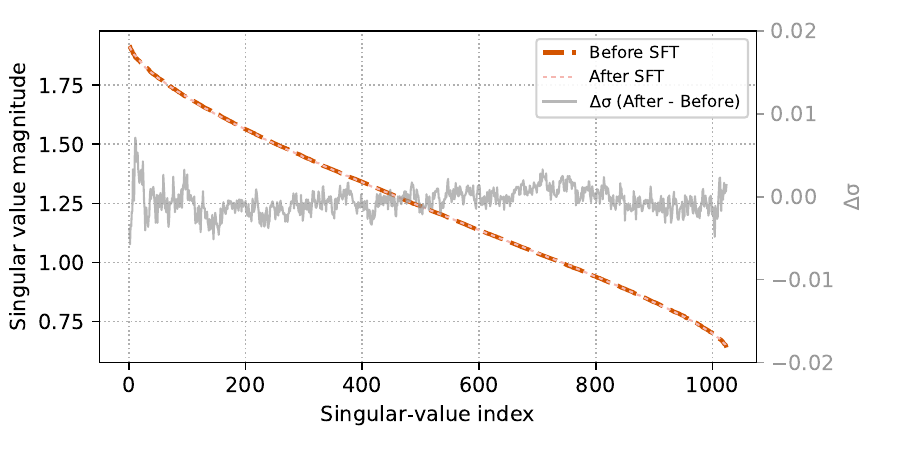}
    \caption{$W_v$ changes during SFT}
    \label{fig:1c}
  \end{subfigure}
  %\vspace{1em} % space between the two rows (adjust as needed)
  % ---- 2nd row -------------------------------------------------
  \begin{subfigure}[b]{0.32\textwidth}
    \includegraphics[width=\linewidth]{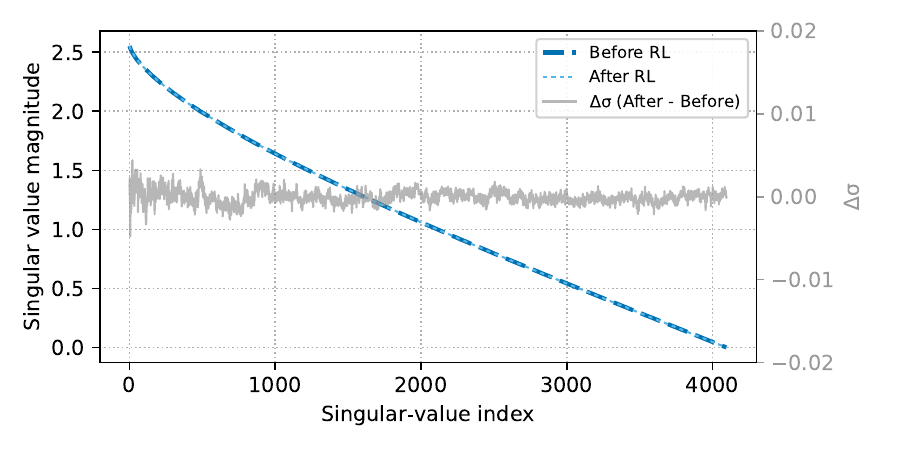}
    \caption{$W_q$ changes during RL}
    \label{fig:2a}
  \end{subfigure}\hfill
  \begin{subfigure}[b]{0.32\textwidth}
    \includegraphics[width=\linewidth]{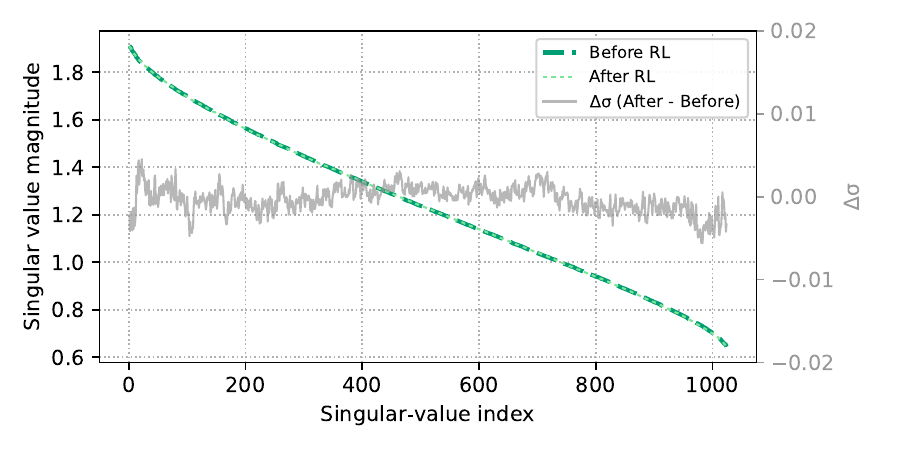}
    \caption{$W_k$ changes during RL}
    \label{fig:2b}
  \end{subfigure}\hfill
  \begin{subfigure}[b]{0.32\textwidth}
    \includegraphics[width=\linewidth]{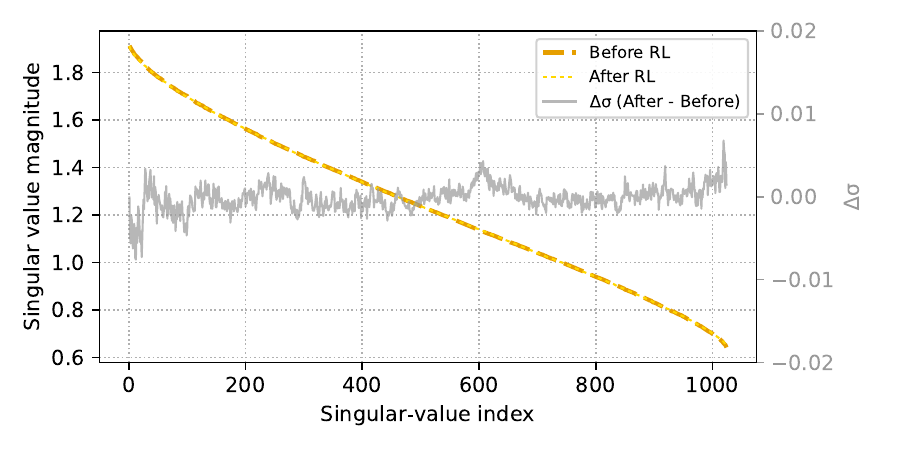}
    \caption{$W_v$ changes during RL}
    \label{fig:2c}
  \end{subfigure}
  \caption{
    Singular value changes in the \texttt{q\_{proj}}, \texttt{k\_proj}, and \texttt{v\_proj} matrices of the first self-attention layer (\texttt{layers[0].self\_attn}) in \texttt{Llama-3.2-11B}. Panels (a)--(c) illustrate the impact of supervised fine-tuning (SFT) on $W_q$, $W_k$, and $W_v$, respectively, while panels (d)--(f) depict the corresponding changes following reinforcement learning (RL). Each panel shows the difference in singular values before and after the respective post-training stage. Note that RL begins from the after SFT checkpoint.
    }
  \label{fig:singular_value_sft_rl}
\end{figure}
\section{Generalization transfer: insights from weight matrices analysis}
\label{sec:param_matrices_analysis}
As discussed in Section 3.2, we observe that during two-stage post-training, the generalization ability reaches peak at early-stage SFT then diminish fastly, while the second-stage RL recovery the generalization ability for a large margin.
To understand the underlying cause of this generalization transfer, we employ Singular Value Decomposition (SVD) to analyze the spectral dynamics of key weight matrices within the transformer architecture.

\subsection{Setup}
Inspired by recent findings from~\cite{staats2025smallsingularvaluesmatter}, which highlight the significance of self-attention matrices in model adaptation, our analysis focuses on two critical sets of matrices:

\textbf{Self-Attention Matrices (\(Q, K, V\))}
The query ($Q$), key ($K$), and value ($V$) weight matrices are the core components of the self-attention mechanism~\cite{vaswani2017attention}. They function by projecting the input embeddings into distinct subspaces to compute attention scores and construct context-aware representations. Analyzing the spectral properties of these matrices is therefore crucial for understanding how the model's fundamental ability to process and weigh relational information is altered during fine-tuning. Shifts in their singular vectors indicate changes in the attention patterns themselves, directly impacting how the model reasons and generalizes from context.

\textbf{Input Embedding and Output Head Matrices}
In models like \texttt{Llama-3.2-11B} and \texttt{Qwen-2.5-7B}, the input embedding (\texttt{embed\_tokens}) and the final output projection (\texttt{lm\_head}) matrices are notably untied. This separation allows us to investigate a key hypothesis: that the drop in generalization during SFT corresponds to a geometric \textbf{divergence} between the input representation space (defined by \texttt{embed\_tokens}) and the output space (defined by \texttt{lm\_head}). Consequently, we hypothesize that the restoration of generalization during RL correlates with their \textbf{re-alignment}. Our spectral analysis is designed to quantify this geometric relationship, offering a novel perspective on the mechanisms of representational change during fine-tuning.

[hangzhan: TODO, we may need to remove most tails of a sentence, it is a clear sign for generative models that always produce some complimentary and unnecessary tails]

% \begin{figure}[ht]
%   \centering
%   % -------- first row --------
%   \begin{subfigure}[t]{0.48\linewidth}
%     \centering
%     \includegraphics[width=\linewidth]{latex/svd_delta-Q-Proj.png}
%     \caption{Q projection}\label{fig:svd_q}
%   \end{subfigure}\hfill
%   \begin{subfigure}[t]{0.48\linewidth}
%     \centering
%     \includegraphics[width=\linewidth]{latex/svd_delta-K-Proj.png}
%     \caption{K projection}\label{fig:svd_k}
%   \end{subfigure}
%   \vspace{0.8em} % vertical space between the two row
%   % -------- second row --------
%   \begin{subfigure}[t]{0.48\linewidth}
%     \centering
%     \includegraphics[width=\linewidth]{latex/svd_delta-V-Proj.png}
%     \caption{V projection}\label{fig:svd_v}
%   \end{subfigure}\hfill
%   \begin{subfigure}[t]{0.48\linewidth}
%     \centering
%     \includegraphics[width=\linewidth]{latex/svd_delta-Down-Proj.png}
%     \caption{Output MLP projection}\label{fig:svd_down}
%   \end{subfigure}
%   \caption{Singular‐value-difference spectra for a random Q, K, V layer and the final MLP layer of Llama-3.2-11B-Base after SFT. The largest (head) and smallest (tail) singular values change the most in every weight matrix.}
%   \label{fig:singular_value_sft}
% \end{figure}

\begin{wrapfigure}{hr}{0.5\linewidth}  % r = right side, width = ~½ \linewidth
 % \vspace{-\baselineskip} 
  \centering
    \includegraphics[width=\linewidth]{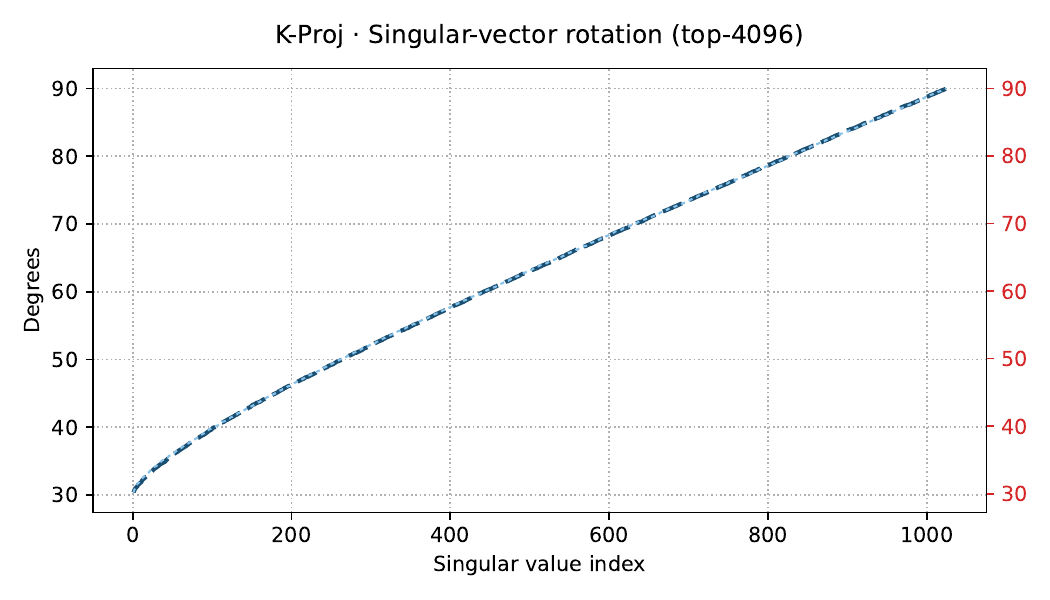}
    \caption{The Principle Angle Analysis of the MLP weight matrix after SFT.}
    \label{fig:rotation_vector}
\end{wrapfigure}

\subsection{Experiment Results}
% (1) For sft, the change mode of q,k,v is similar, change slightly on the head and tail of singular value index, the reason is that the head singular value is largest and the delta will be relatively larger. (2) observe similar phonement on rl. (3) overall "energy" of the learned transformations are largely preserved from the pre-trained model for both sft and rl. explain energy regards to singular value. (4) the model performance during training has little correlation with singular value.

\textbf{Singular Value Dynamics during SFT and RL.} 
To investigate how SFT and RL reshape the spectral structure of the model , we analyze the singular values (SVs) of the \textsc{Q/K/V} and their differences ($\Delta\sigma_i = \sigma_i^{\text{after}} - \sigma_i^{\text{before}}$) before/after different training stage. The results are shown in the Figure~\ref{fig:singular_value_sft_rl}. We found that: \textbf{the singular value distributions of the $Q, K, V$ matrices remain remarkably stable after both SFT and RL across all experiments}. For both fine-tuning paradigms, the plots of singular values "Before" and "After" training are nearly indistinguishable. The change in singular values, represented by $\Delta\sigma$, fluctuates from 0 to 0.005, as a low-magnitude, zero-centered noisy signal across the entire singular-value index for all Q, K, and V projections. This indicates that the fine-tuning process does not systematically amplify or diminish specific singular values. 

While there are slight variations for both SFT and RL, the pattern of change in Q, K, and V matrices is broadly similar, consisting of minor adjustments across the spectrum rather than significant alterations concentrated in any particular region, such as the head (largest values) or tail (smallest values). A similar phenomenon is observed for RL, where the modifications are also subtle and distributed.

\textbf{Preservation of Transformation Energy.}
As the singular values $\sigma_i$ of a matrix quantify the magnitude of its "stretching" effect along its principal axes~\cite{}. The overall "energy" of the transformation, often related to the Frobenius norm squared ($||W||_F^2 = \sum_i \sigma_i^2$)~\cite{}, represents the total amplification capacity of the matrix. Given that the singular value spectra are almost unchanged after both SFT and RL, it follows that the "energy" of these learned projection matrices is largely preserved from the pre-trained model. This suggests that fine-tuning primarily learns to re-orient the transformation in the high-dimensional weight space rather than altering its overall amplification characteristics or effective rank.
%As illustrated in Figure~\mathbb{R}ef{fig:1a}, \mathbb{R}ef{fig:1b} and \mathbb{R}ef{fig:1c}, we observe that SFT induces subtle but highly targeted modifications to the model's weight matrices. For best visulization, we only visualize the singular value of Q,K,V and MLP project for random picked layer(layer 0 in Figure~\mathbb{R}ef{fig:singular_value_sft_rl}). More singular value spectra are shown in Appendix. Overall, the magnitudes of the singular values remain remarkably stable throughout SFT. This indicates that the intrinsic rank and the overall "energy" of the learned transformations are largely preserved from the pre-trained model.

Limitations Or open questions

\textbf{Rotation of singular vectors matters.} %Compared with consistent singular value of weight matrices, we observe that singular vector rotate during SFT as illustrated in Figure~\mathbb{R}ef{fig:rotation_vector}. The small angles for low-index vectors indicate that core, dominant features are largely preserved. Conversely, the angles approach 90 degrees for high-index vectors, signifying a near-complete redefinition of the fine-grained features. Our results suggest that the primary mechanism of adaptation during SFT is the rotation of singular vectors. The model is not changing how much importance it places on its learned features, but rather it is redefining what those features are. This re-orientation of the model's representational axes is a more subtle and efficient form of adaptation, and it directly explains the two-stage generalization pattern we observe. \sitao{how does it explains the two-stage generalization pattern?} 
As described in Section 2.2, principal angles measure the 'tilt' between corresponding singular vectors of two matrices: 0 means the vectors are identical, 90 means they are orthogonal. We use principle angle of singular vector before and after SFT or RL to analyze the rotation characteristic during SFT and RL.
For illustration, we plot the principal angle spectrum of the layer-0 up$_\text{proj}$ matrix in Figure~\ref{fig:rotation_vector}; the corresponding spectra for the $K$ and $V$ matrices, as well as for deeper layers, follow the same trend and are reported in Appendix. In every case, the curves for supervised SFT and RL lie on top of each other: angles start around 25-30 for the leading singular vectors and rise smoothly toward 90 in the tail. This tight overlap implies that the network adapts primarily by rotating its singular vectors in virtually the same way for SFT and RL while leaving the singular values largely intact. In other words, both training stages preserve core, low index features but progressively redefine higher index, fine-grained directions.

While our empirical results in the next section show that the rotation during SFT and RL correlates with OOD performance, yet the exact nature of the rotation pattern shared by SFT and RL remains unresolved. Understanding why the two optimisation regimes converge on the same rotation profile is an open question that we will investigate in future work.

%\textbf{Early SFT}
% Generalization Recovery via Spectral Intervention
\subsection{Singular Vector Directional Re-alignment of SFT Weights}
To validate our hypothesis that SFT-induced vector rotation is the primary cause of generalization loss, we conducted experiments to reverse these rotations. We replaced the singular vectors directions(U and V matrices) of the post-SFT weights with those from the original pre-trained model, effectively "un-doing" the directional changes while keeping the SFT-learned singular values.

\begin{figure}[tp]        % h = “here”, t = top, b = bottom, p = page of floats
  % --- first subfigure ------------------------------------
  \begin{subfigure}[t]{0.46\linewidth}
    \centering
    \includegraphics[width=\linewidth]{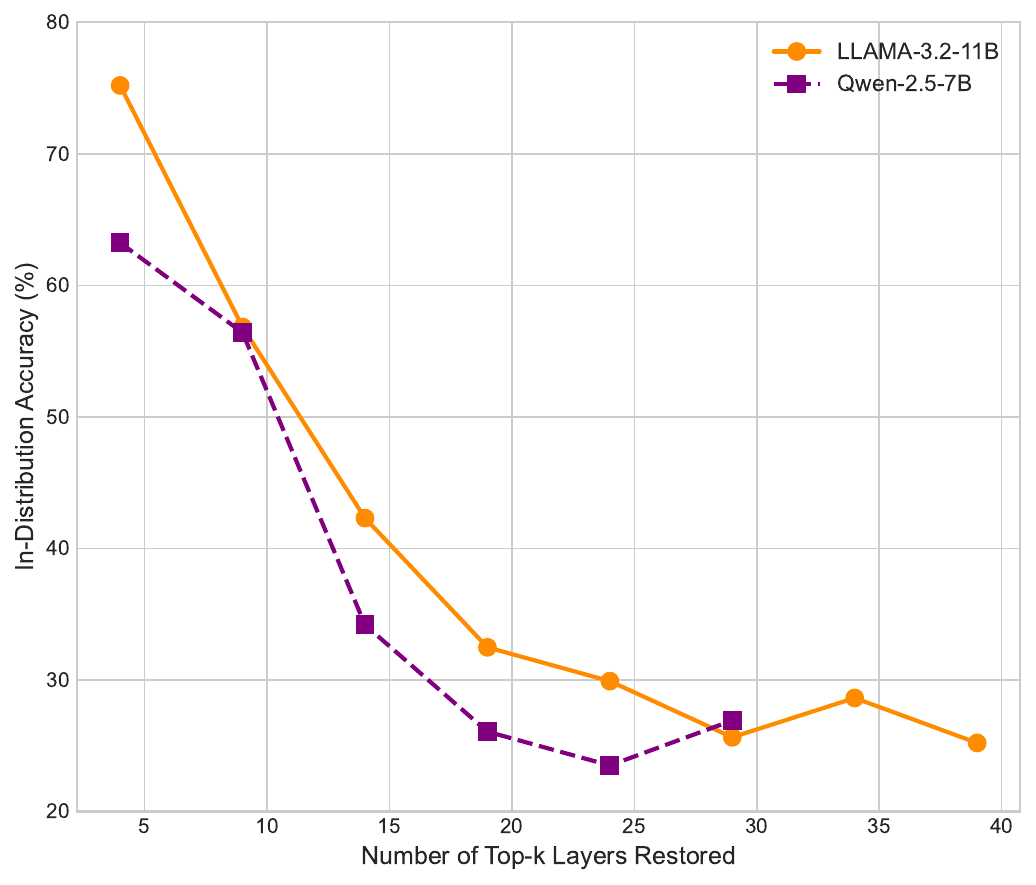} % first-image.pdf/png/eps…
    %\caption{First image caption}
    \label{fig:first}
  \end{subfigure}
  \hfill                           % horizontal space between subfigures
  % --- second subfigure -----------------------------------
  \begin{subfigure}[t]{0.46\linewidth}
    \centering
    \includegraphics[width=\linewidth]{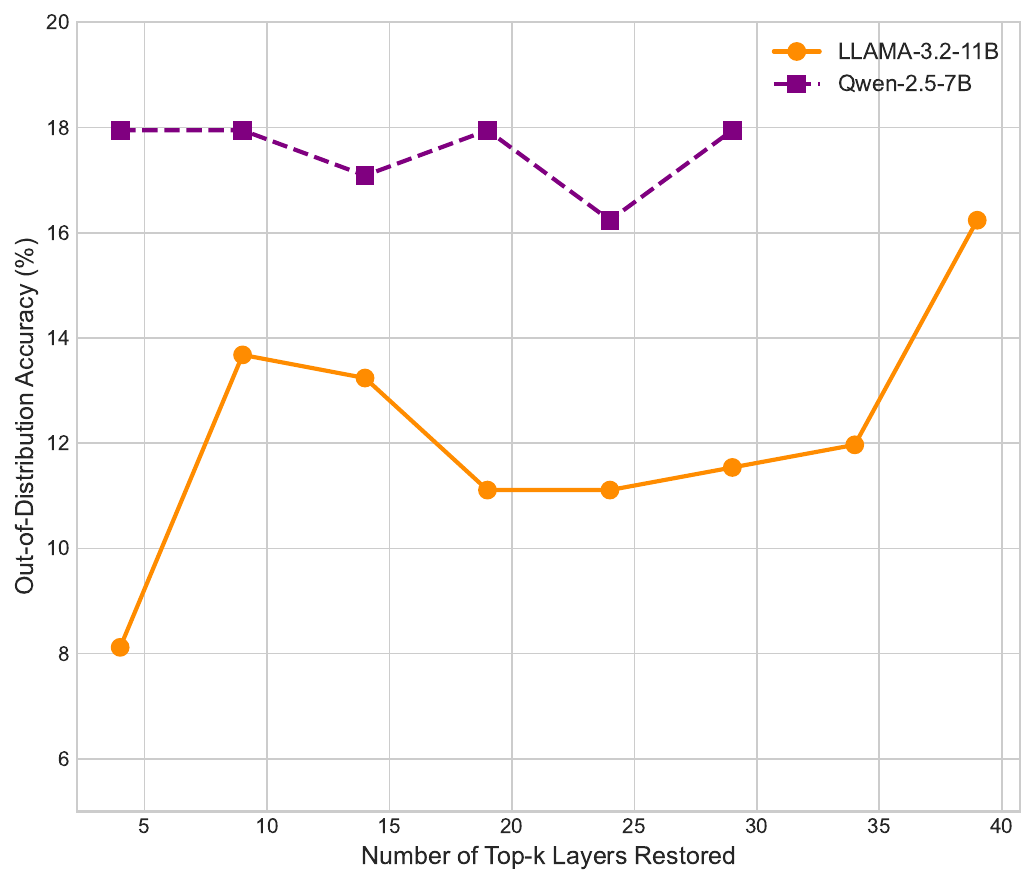}
    %\caption{Second image caption}
    \label{fig:second}
  \end{subfigure}
  \caption{Impact of restoring top-k layers' singular directions on Out-of-Distribution (OOD) and In-Distribution (ID) accuracy.}
  \label{fig:recover_layer_rank}
\end{figure}

\begin{figure}[t]        % h = “here”, t = top, b = bottom, p = page of floats
  % --- first subfigure ------------------------------------
  \begin{subfigure}[t]{0.46\linewidth}
    \centering
    \includegraphics[width=\linewidth]{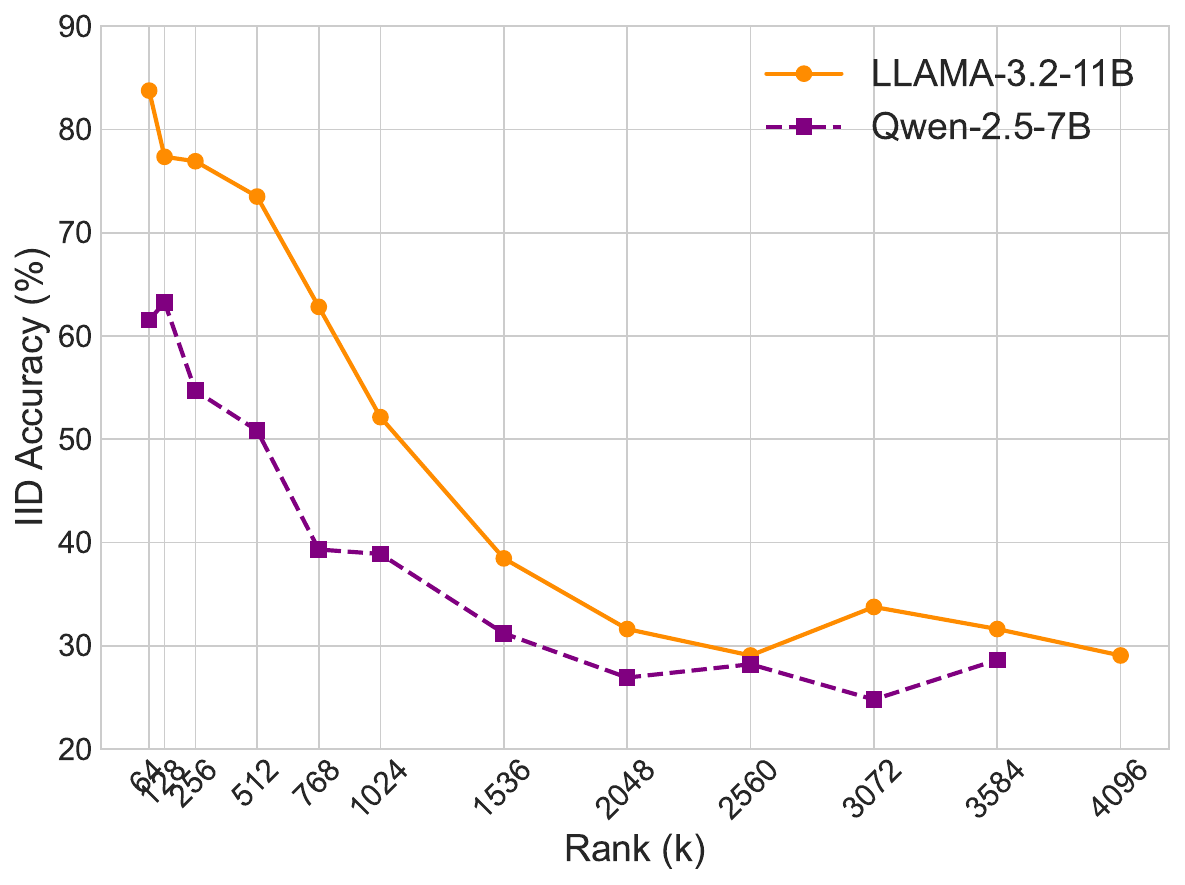} % first-image.pdf/png/eps…
    %\caption{First image caption}
    \label{fig:first}
  \end{subfigure}
  \hfill                           % horizontal space between subfigures
  % --- second subfigure -----------------------------------
  \begin{subfigure}[t]{0.46\linewidth}
    \centering
    \includegraphics[width=\linewidth]{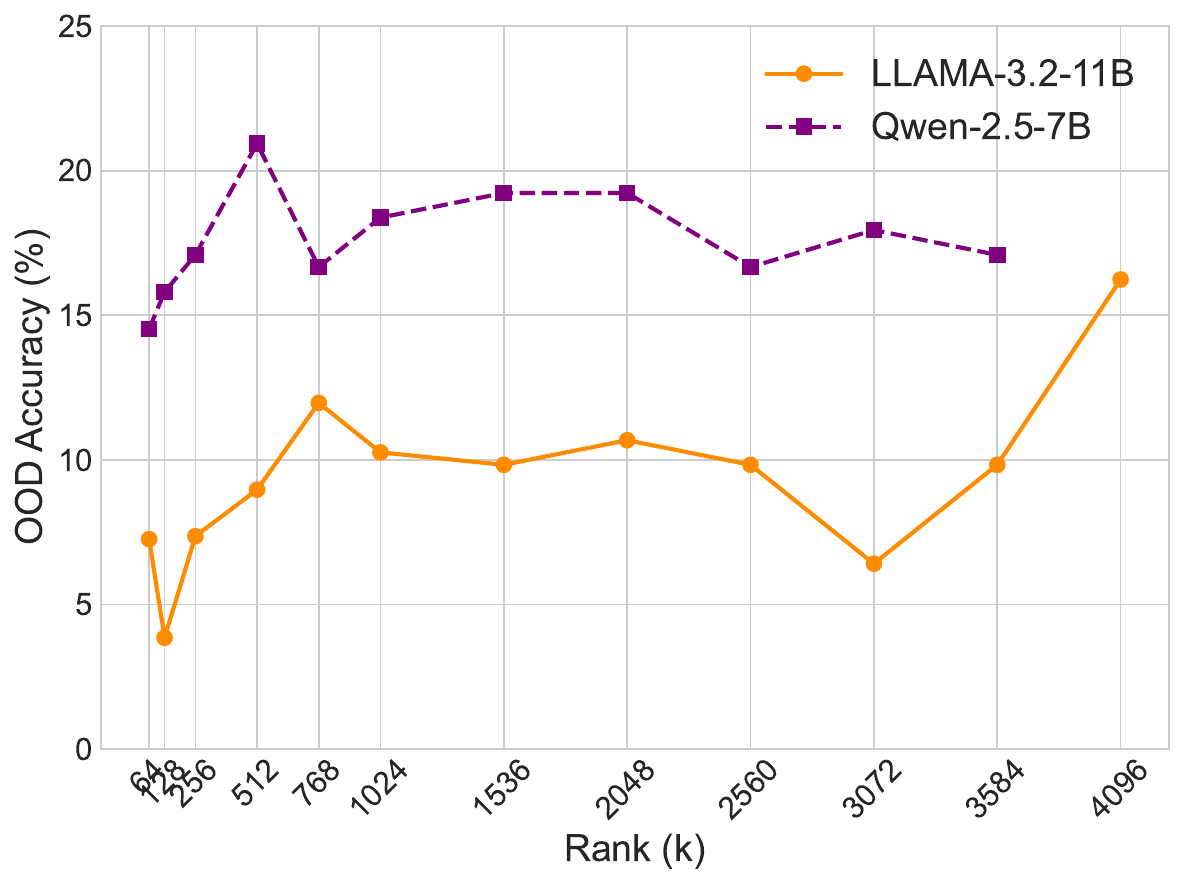}
    %\caption{Second image caption}
    \label{fig:second}
  \end{subfigure}
  \caption{Impact of restoring top-k ranks' singular directions on Out-of-Distribution (OOD) and In-Distribution (IID) accuracy.}
  \label{fig:rank_restore}
\end{figure}

\textbf{Layer-wise Reversing Rotations.} To pinpoint where SFT-induced functional misalignment occurs, we conducted a granular analysis by selectively restoring the singular vector directions of different top-K layers. We discovered that: (1) restoring the intermediate layers (\eg{} layers 10-20 for LLAMA and 10-15 for Qwen) was most detrimental to in-distribution (IID) performance, causing a significant drop in task accuracy. This suggests that SFT primarily concentrates task-specific knowledge and adaptations within the central block of the model;  (2) restoring the head (\eg{} 0-9) and tail (\eg{} 20+) layers had a more pronounced effect on out-of-distribution (OOD) performance. Reversing the geometric changes in these outer layers was most effective at recovering the model's general reasoning capabilities, often at a lesser cost to IID accuracy. This strongly indicates that the model's general functional alignment is maintained by the shallower and deeper layers, while the intermediate layers are repurposed to specialize during fine-tuning.
%As illustrated in right part of Figure~\mathbb{R}ef{fig:recover_layer_rank}, re-aligning just the first 10 layers boosts OOD accuracy from a baseline of ~4.7\% to a peak of ~13.7\%. While adding more layers initially causes a slight dip, recovering all 40 layers results in the highest accuracy of over 16\%, nearly restoring the model to its peak Inter\_SFT performance. This demonstrates that the detrimental rotations are distributed across the entire model, and a full, layer-wise re-alignment is necessary to maximally recover lost generalization.

\begin{wrapfigure}{hr}{0.5\linewidth}  % r = right side, width = ~½ \linewidth
  %\vspace{-\baselineskip}              % fine-tune vertical alignment (optional)
  \centering
  \includegraphics[width=\linewidth]{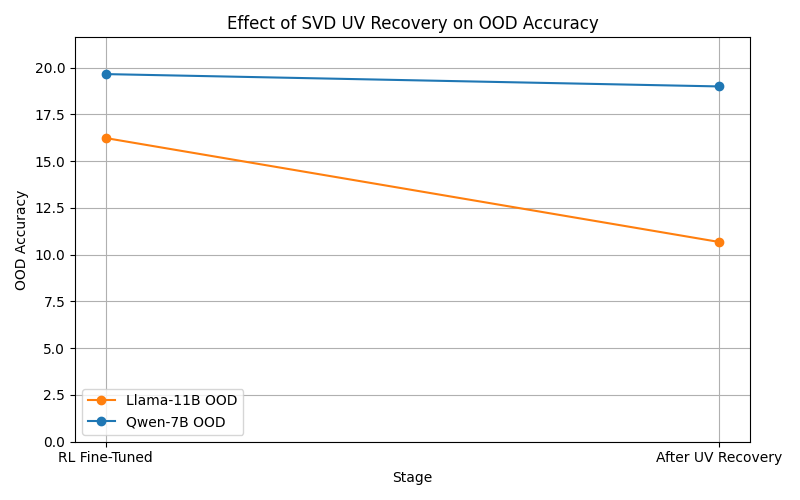}
  \caption{Effect of Imposing SFT-Learned Feature Directions on RL-Tuned Models.}
    \label{fig:rl_to_sft}
\end{wrapfigure}

\textbf{Rank-wise Reversing Rotations.} We further analysis which components within each layer are most critical for this recovery by reversing rank-wise weight matrices rotations. From the results as plotted in Figure~\ref{fig:rank_restore}, we concluded that (1)task-specific knowledge learned during SFT is broadly distributed across the singular spectrum as shown in IID performance; (2) generalizable, foundational knowledge is primarily encoded in the top-rank singular directions. As seen in the right panel, restoring these principal components has the most significant positive impact on OOD accuracy. For Qwen, performance peaks after restoring approximately 512 directions and then stabilizes, suggesting that the core general reasoning capabilities are concentrated in this high-rank subspace. LLAMA shows a similar, albeit noisier, trend where OOD performance generally improves as more directions are restored.
%In the left part of Figure~\mathbb{R}ef{fig:recover_layer_rank}, recovering just the top 768 components (out of 4096) is sufficient to achieve a significant accuracy peak of ~11.9\%.While recovering only the top components provides a substantial boost, the highest performance (~16.2\%) is achieved only when all 4096 components are recovered. This indicates a dual mechanism: the most significant gains in generalization come from re-aligning the core, dominant features (the "head" of the spectrum), but full recovery requires correcting the incoherent rotations across the entire feature space, including the fine-grained details at the "tail." 

\textbf{Causal Validation of RL's Rotational Recovery} By taking the high-performing RL-tuned model and forcing its feature space to adopt the geometric orientation of the poorly-generalizing SFT model, we effectively reverse the benefits of RL. As shown in Figure~\ref{fig:rl_to_sft}, OOD accuracy plummets from 16.2\% down to 10.6\%, a loss of over a third of its generalization advantage on Llama-3.2-11B-Base. This shows unequivocally that the specific vector directions found by RL are essential to its success and are fundamentally different from the directions settled upon by SFT. The result for the Qwen-2.5-7B-Base model, while less dramatic, still drops from 19.8\% to 19.0\%. This suggests that while RL still finds a superior geometric arrangement for Qwen, the feature directions may be closer to those of its SFT counterpart compared to Llama.

\section{Related Work}
\label{sec:related_work}
The optimal integration of Supervised Fine-Tuning (SFT) and Reinforcement Learning (RL) for aligning language models remains an open question. \cite{ma2025learning} provide evidence supporting this separation, reporting that RL is most effective for enhancing robustness on low- to medium-difficulty tasks, while SFT's structured demonstrations yield superior gains on complex reasoning problems. This suggests a division of labor where each method addresses a different facet of model improvement.
However, the efficacy of this sequential pipeline is not universally accepted. A key challenge, identified by \cite{zhang2025bread}, is that SFT data often contains implicit "reasoning leaps" that imitation learning struggles to internalize. This deficiency subsequently hinders the RL phase by limiting its ability to explore and generate high-reward trajectories. Motivated by such limitations, a growing body of work argues against a rigid SFT-then-RL separation, instead proposing to integrate SFT and RL into a single, unified training stage \cite{liu2025uft, huang2025blending}.
Further analysis reveals a fundamental tension between the two learning objectives. \cite{fu2025srft} find that the SFT loss pulls the model policy substantially away from its base initialization, whereas the KL-regularized RL objective acts as a countervailing force, constraining the policy to remain close to the base model. This highlights an underlying conflict that can impede optimization. Our work is situated within this context, aiming to reconcile this tension.

Previous studies \cite{chu2025sftmemorizesrlgeneralizes} compare the generalization between supervised fine-tuning and reinforcement learning methods, claiming that RL methods based on the PPO algorithm exhibit better generalization capability compared to supervised fine-tuning, which tends to memorize training data. Another line of research \cite{shuttleworth2024loravsfinetuningillusion} indicates that LoRA can achieve performance comparable to fully fine-tuned models but struggles with robustness due to the presence of intruder dimensions caused by LoRA fine-tuning.

Zheng et al. \cite{zheng2025spuriousforgettingcontinuallearning} explore the phenomenon of ``spurious forgetting'' in continual learning scenarios, suggesting that performance degradation often reflects reduced task alignment rather than actual loss of knowledge. Their findings underline the importance of distinguishing task alignment from knowledge retention to improve continual learning strategies. Logic-RL \cite{xie2025logicrlunleashingllmreasoning} demonstrates the effectiveness of rule-based RL in enhancing LLM reasoning skills, showing significant generalization to complex benchmarks such as AIME and AMC after training on synthetic logic puzzles.

Swamy et al. \cite{swamy2025roadsleadlikelihoodvalue} investigate the rationale behind the effectiveness of RL-based fine-tuning despite its complexity compared to offline likelihood estimation, emphasizing RL's advantage in scenarios involving a generation-verification gap. Misu et al. \cite{Misu_2024} highlight the integration of formal verification methods in improving LLM-generated code correctness. Lastly, Ren and Sutherland \cite{ren2025learningdynamicsllmfinetuning} analyze learning dynamics during LLM fine-tuning, providing insights into why certain forms of hallucinations persist post-fine-tuning and proposing frameworks for understanding and enhancing model alignment.

While these studies corroborate the view that RL excels at generalization beyond the direct training set, our work aims to dive deeper into understanding \emph{why}---and in which specific parameter regimes---RL training leads to stronger OOD capabilities. In particular, by focusing on the 24-point card game and a math reasoning task, we highlight the potential gains and pitfalls of RL with respect to reward hacking when no careful initialization (\ie{} via SFT) is employed.

\section{Conclusion and Future Work}
\label{sec:conclusion_future_work}
Conclusions and Future Directions:
Based on our analysis, we derive three primary conclusions:
\begin{itemize}
\item PCA and singular values undergo minimal overall change during supervised fine-tuning.

\item Singular value adjustments primarily occur at the spectrum's extremes (head and tail), potentially implicating these extremes in affecting generalization.

\item RL fine-tuning does not universally enhance OOD generalization but is effective in countering moderate overfitting induced by intermediate-stage supervised fine-tuning. Once severe overfitting occurs, RL alone cannot restore generalization capabilities.
\end{itemize}

To substantiate conclusions (2) and (3), we are currently undertaking additional experiments. Specifically, we aim to isolate the head and tail singular values to ascertain their exact role in deteriorating model generalization. Moreover, we plan to perform RL fine-tuning at multiple intermediate and late checkpoints across varied reasoning tasks (such as math reasoning) to comprehensively validate the robustness and universality of our findings.

We plan to expand this line of investigation by examining detailed weight trajectories, focusing on which layers drive OOD improvements, and applying the same methodology to different tasks (\eg{} advanced math and code-generation tasks). These ongoing experiments are critical to establishing the precise mechanisms through which RL and SFT affect LLM generalization, guiding better fine-tuning strategies.

%TODO sweet point signal
%\sitao{Layer-wise analysis. comparison with other small singular value analysis of LLM finetuning}
\newpage
\bibliographystyle{plainnat}
\bibliography{main}

\newpage

\appendix
\section{Prove}

2 layers linear MLP.
To prove if it is trivial. and how to prove the rotation is more important than the update of the singular values.
TODO connection between QKV and sub-chain
\begin{theorem}  1 (Rotation Matters)
Suppose we have input $X \in \mathbb{R}^{ d \times n}$ and each column $X_{:,i}$ is a token,  for a transformer with components $Q, K, V$
    \begin{equation}
        Q= W_Q X,\ K = W_K X,\ V=  W_V X.
    \end{equation}
Suppose we have a sub-chain in a transformer
    \begin{align}
          X \;\xrightarrow{\,W_1\,}\; H
          \;\xrightarrow{\,W_2\,}\;
          Y = W_2 W_1 X,
    \end{align}
Then, suppose  $W=[W_1||W_2]$. For the loss function with weight decay
    \begin{equation}
\mathcal{L}(\mathbf{W}) = \frac{1}{n} \sum_{i=1}^{n} (Y_i - \hat{Y}_i)^2 + \lambda \|\mathbf{W}\|_2^2
    \end{equation}
Suppose $A$ is skew-symmetric and $W_{1}$  and $W_2$ are bounded with finite norm, learning rate $\eta \ll  1$, and after one gradient descent step, suppose we have the following update
\begin{equation}
  \label{eq:update}
  \Delta W_1 = +\eta W_1 A,
  \qquad
  \Delta W_2 = -\eta A W_2,
\end{equation}
%Suppose $A$ is skew-symmetric and $W_{1}$  and $W_2$ are bounded with finite norm, after one step gradient descent,
then, $\|\Delta W\|_F^2 = \mathcal O(\eta^2)$, \ie{}\textbf{two orders smaller}
than a direct change of singular values
$\|\Delta\Sigma_1\|_F^2 = \mathcal O(\eta), \|\Delta\Sigma_2\|_F^2 = \mathcal O(\eta)$, where $\Delta\Sigma_1$ and  $\Delta\Sigma_2$ are the singular values of $\|\Delta W_1\|$ and $\|\Delta W_2\|$.

%\(
%    \mathcal L + \lambda\!\sum\|W\|_F^2,
%\)
% the optimiser naturally drifts along this
% \emph{orthogonal-gauge manifold} rather than paying the higher cost of altering $\Sigma$.

    \end{theorem}

\subsection{Orthogonal--rotation gauge: why fine--tuning prefers to rotate rather than altering singular values}
\paragraph{Notation and standing assumptions.}
\begin{proof}

\begin{description}
  \item[Block sub--chain]
        We isolate within one Transformer block the \emph{two layer linear
        chain}
        % \[
        %   X \;\xrightarrow{\,W_1\,}\; H
        %   \;\xrightarrow{\,W_2\,}\;
        %   Y = W_2 W_1 x,
        % \]
        % where $W_1 \in \mathbb{R}^{d_{\mathrm{mid}}\times d}$ and
        % $W_2 \in \mathbb{R}^{d\times d_{\mathrm{mid}}}$.
        % This sub-chain covers
        % \begin{itemize}\setlength\itemsep{0pt}
        %   \item \textbf{Self-attention branch:}\;
        % \(W_1 = W_{\mathbf K\!/\!\mathbf V}\,,\;
        %   W_2 = W_{\mathbf O}\),
        % with \(d_{\text{mid}} = n_h d_h = 32\times32=1024\).
        %   \item \textbf{MLP branch:}\;
        %         \(W_1 = W_{\text{up}},\;
        %           W_2 = W_{\text{down}},\;
        %           d_{\text{mid}} = d_{\text{ff}} = 14\,336\).
        % \end{itemize}
  \item[Learning-rate symbol]
        $\eta \ll 1$ is the step size of one gradient update
        (e.g.\ $\eta\!\sim\!10^{-4}$ in fine-tuning).

  \item[Skew matrix $A$]
        $A\in\mathbb{R}^{d_{\mathrm{mid}}\times d_{\mathrm{mid}}}$ with
        $A^\top=-A$.  
        Such $A$ generates an \emph{infinitesimal rotation}.

  \item[First order in $\eta$]
        We keep terms linear in $\eta$ and absorb
        quadratic and higher terms into the symbol
        $\mathcal O(\eta^{2})$.
        Numerically, with $\eta\!\sim\!10^{-4}$,
        $\eta^{2}\!<\!10^{-8}$ is negligible.

  \item[Orthogonal increment]
        \[
          R_\eta := I+\eta A,
          \qquad
          R_\eta^{-1}= I-\eta A + \mathcal O(\eta^{2}),
          \qquad
          R_\eta^\top R_\eta = I + \mathcal O(\eta^{2}).
        \]
        Thus $R_\eta$ is \emph{orthogonal to first order}.
        Over many steps the product
        $\prod_t (I+\eta A_t) = \exp(\sum_t\eta A_t)$
        lies exactly in the orthogonal group $O(d_{\mathrm{mid}})$.
  \item[$W^{\text{fin}}$]
        The final W.     
  \item[$d_{\mathrm{mid}} (bottleneck width)$]
        The inner dimension of the sub-chain.
        Typical values in LLaMA-3.2 are  
        $d_{\mathrm{mid}} = n_h d_h \;(=\;1024)$ for attention,  
        or $d_{\mathrm{mid}} = d_{\mathrm{ff}}\;(=\;14\,336)$ for the MLP.      
  \item[$d_{\mathrm{model}}$]
        Model width (hidden size), \eg{} 4096 in LLaMA-3.2.
  
  \item[$d_{\mathrm{ff}}$]
        Feed-forward width; usually larger than $d_{\mathrm{model}}$.

\end{description}

\paragraph{Setup.}
Given the linear sub-chain inside one Transformer block
\[
  x \;\longmapsto\; W_1 x
  \;\longmapsto\; W_2 W_1 x,
  \qquad
  W_1 \in \mathbb{R}^{d_{\mathrm{mid}}\times d},\;
  W_2 \in \mathbb{R}^{d\times d_{\mathrm{mid}}}.
\]
For self-attention we have $(W_1,W_2)=(k/v_\text{proj},o_\text{proj})$ with
$d_{\mathrm{mid}} = n_h d_h$;
for the MLP branch $(W_1,W_2)=(up_\text{proj},down_\text{proj})$ with
$d_{\mathrm{mid}}=d_{\mathrm{ff}}$.

\vspace{3pt}
\paragraph{A single gradient step.}
Let $A \in \mathbb{R}^{d_{\mathrm{mid}}\times d_{\mathrm{mid}}}$ be \emph{skew-symmetric}
($A^\top=-A$) and $\eta\ll1$ a step size.
If the optimiser chooses the update
\begin{equation}
  \label{eq:update}
  \Delta W_1 = +\eta W_1 A,
  \qquad
  \Delta W_2 = -\eta A W_2,
\end{equation}
then the new weights are
\[
  W_1' = W_1(I+\eta A)=W_1 R_\eta,
  \quad
  W_2' = (I-\eta A)W_2 = R_\eta^{-1} W_2,
  \qquad
  R_\eta := I+\eta A \in O(d_{\mathrm{mid}})+\mathcal O(\eta^2).
\]

\paragraph{No change to the forward function.}
To first order in $\eta$
\[
  W_2' W_1' \;=\;
  (R_\eta^{-1} W_2)(W_1 R_\eta)
  \;=\; W_2 W_1 + \mathcal O(\eta^2),
\]
so \emph{functional mapping} of the block is preserved.

\paragraph{Infinitesimal parameter cost.}
The squared Frobenius change is
\[
  \begin{aligned}
  \|\Delta W_1\|_F^2 + \|\Delta W_2\|_F^2
  &= \eta^2 \bigl(\|W_1 A\|_F^2 + \|A W_2\|_F^2 \bigr) \\
  &= \eta^2 \trace\bigl(A^\top W_1^\top W_1 A + A W_2 W_2^\top A^\top \bigr) .
  \end{aligned}
\]
Because $A$ is skew-symmetric and $W_{1,2}$ already have finite norm,
the cost scales as $\mathcal O(\eta^2)$ -- \textbf{two orders smaller}
than a direct change of singular values
($\Delta\Sigma = \mathcal O(\eta)$ gives $\|\Delta W\|_F^2=\mathcal O(\eta)$).

\paragraph{Accumulating many small steps.}
Over $T$ iterations the product of
$R_\eta = I+\eta A_t$ exponentiates to a \emph{strictly orthogonal} matrix
\[
  R \;=\; \prod_{t=1}^{T} (I+\eta A_t)
    \;=\; \exp\!\Bigl( \textstyle\sum_{t=1}^{T}\eta A_t \Bigr)
    \in O(d_{\mathrm{mid}}).
\]
Consequently
\[
  W_1^{\text{fin}} = W_1 R,
  \qquad
  W_2^{\text{fin}} = R^{-1} W_2,
\]
while the composite map $W_2^{\text{fin}} W_1^{\text{fin}}$ is (almost) unchanged.

\paragraph{Effect on SVD.}
Write the truncated SVD of $W_1$ as $W_1 = U \Sigma V^\top$.
Right-multiplying by \emph{orthogonal} $R$ leaves $\Sigma$ \emph{invariant}
but rotates the right singular vectors:
\[
   W_1 R
   \;=\;
   U \Sigma (V^\top R)
   \quad\Longrightarrow\quad
   \Sigma'=\Sigma,\;
   V' = VR.
\]
Analogously for $W_2$ the left singular vectors $U$ rotate by $R^{-1}$.
Hence fine-tuning produces the empirical signature you observe:

\begin{itemize}
  \item \textbf{Singular values stay nearly fixed.}
  \item \textbf{Left / right singular vectors rotate by large angles}
\end{itemize}
\end{proof}

\paragraph{Why the optimiser prefers rotation.}
Updating via \eqref{eq:update} simultaneously
\emph{(i)} preserves the forward loss,
\emph{(ii)} incurs only $\mathcal O(\eta^2)$ weight-decay penalty in SFT, KL penalty and gradient clip in RL,
\emph{(iii)} avoids numerical instabilities (orthogonal transforms keep vector norms).
Thus, under a typical objective
\(
    \mathcal L + \lambda\!\sum\|W\|_F^2,
\)
the optimiser naturally drifts along this
\emph{orthogonal-gauge manifold} rather than paying the higher cost of altering $\Sigma$.

Note: if $\eta >=1$, Our prove doesn't hold.

The toy code below can simply demonstrate this process:

\begin{python}
import torch, math
    torch.set_printoptions(precision=5)

    W1 = torch.eye(2)
    W2 = torch.tensor([[1., 0.],
                       [0., 0.5]])

    theta = math.radians(10)
    R_true = torch.tensor([[math.cos(theta), -math.sin(theta)],
                           [math.sin(theta), math.cos(theta)]])

    W1_rot = W1 @ R_true
    W2_rot = R_true.T @ W2

    def procrustes_R(A, B):
        U, _, Vt = torch.linalg.svd(A.T @ B, full_matrices=False)
        return U @ Vt

    R_hat = procrustes_R(W1, W1_rot)

    print("True R:", R_true)
    print("Estimated R:", R_hat)
    print("||R_hat - R_true||_F =", (R_hat - R_true).norm())

    W1_aligned = W1_rot @ R_hat.T
    W2_aligned = R_hat @ W2_rot

    print("delta norm after rotation =", (W1_aligned - W1).norm() ** 2
          + (W2_aligned - W2).norm() ** 2)

    x = torch.randn(5, 2)
    out_base = x @ W1 @ W2
    out_align = x @ W1_aligned @ W2_aligned
    print("max |out| difference =", (out_align - out_base).abs().max())

\end{python}
% \section{Example Appendix}
% \label{sec:appendix}

% This is an appendix.

\section{Experiment Settings}
\label{app:setting}
All our experiments are implemented on 4 H100 GPUs. For SFT, the learning rate is 1e-6, a mini batch size of 64, and a roll
%with a constant learning rate of 5e-7, a mini batch size (number of rollouts seen before a gradient update) of 128, and a rollout batch size (number of prompts we rollout at the same time) of 64. For each prompt, we collect 16 rollouts to compute advantages for GRPO update. We use a sampling temperature τ = 1. We do not apply KL divergence loss or entropy loss in our training.

\label{app:examples}

\end{document}